\documentclass[prl, twocolumn, showpacs, superscriptaddress]{revtex4-1} 
\usepackage{amsmath}   
\usepackage{amsfonts} 
\usepackage{graphicx}  
\usepackage{subfigure}

\begin{document}
\bibliographystyle{unsrt}
\title{Spectral Ergodicity in Deep Learning Architectures 
      via Surrogate Random Matrices}
\author{Mehmet S\"uzen}
\affiliation{INM-6, Forschungszentrum Juelich and Department of Neurobiology, 
         Ludwig-Maximilians-Universit\"at M\"unchen, Germany} 
\email{mehmet.suzen@physics.org}
\author{Cornelius Weber}
\affiliation{Department of Informatics, Faculty of Mathematics and 
            Natural Sciences, University of Hamburg, Germany}
\email{weber@informatik.uni-hamburg.de}
\author{Joan J. Cerd\`{a}}
\affiliation{Department of Physics, University of the Balearic Islands, 
             Crta. de Valldemossa, km 7.5, 07122 Palma (Illes Balears), Spain}
\email{jj.cerda@uib.cat}
\date{\today}
\begin{abstract}
In this work a novel method to quantify spectral ergodicity for random matrices
is presented. The new methodology combines approaches rooted in the metrics of 
Thirumalai-Mountain (TM) and Kullbach-Leibler (KL) divergence. The method is 
applied to a general study of deep and recurrent neural networks via the analysis 
of random matrix ensembles mimicking typical weight matrices of those systems. 
In particular, we examine circular random matrix ensembles: circular unitary 
ensemble (CUE), circular orthogonal ensemble (COE), and circular symplectic 
ensemble (CSE). Eigenvalue spectra and spectral ergodicity are computed for 
those ensembles as a function of network size. It is observed that as the 
matrix size increases the level of spectral ergodicity of the ensemble 
rises, i.e., the eigenvalue spectra obtained for a single realisation 
at random from the ensemble is closer to the spectra obtained averaging 
over the whole ensemble. Based on previous results we conjecture that success 
of deep learning architectures is strongly bound to the concept of spectral 
ergodicity. The method to compute spectral ergodicity proposed in this work 
could be used to optimize the size and architecture of deep as well 
as recurrent neural networks.
\end{abstract}


\maketitle

Applications of random matrices appear in a wide range of 
fields \cite{mehta2004random, edelman2005a, Mezzadri2005,
             forrester2010, couillet2011, bai2014}. 
Characterising statistical properties of different random matrix ensembles 
plays a critical role in understanding the nature of physical models they 
represent. For example in neuroscience, neuronal dynamics can be encoded 
by means of a synaptic connectivity matrix in different network 
architectures \cite{rajan2006b, aljadeff15a, parisi2015a, 
                    ahmadian2015a, rajan2016a}, such as 
deep learning architectures \cite{lecun2015a,bengio2009a}, possibly with 
dropout \cite{srivastava2014a}. Furthermore, transition matrices in 
stochastic materials simulations \cite{cerda2005a, suezen2014a} in 
discrete space appear as a realisation of a random matrix ensemble.

One widely studied statistical property of such matrices lies in 
spectral analysis. The eigenvalue spectrum entails information regarding both 
structure and dynamics. For example, spectral radius of weight matrices 
in recurrent neural networks influence the learning dynamics, i.e., training 
\cite{pascanu13a}. Ergodic properties are not much investigated in this 
context, though they have been studied in depth in the case of quantum 
systems \cite{haake2013a} where energy spectra are a main source of 
information that stems from those systems.

The concept of ergodicity appears in statistical mechanics in that the
time average of a physical dynamics is equal to its ensemble 
average \cite{peters13a,  suezen2014a}. The definition is not uniform 
in the literature \cite{suezen2014a}. For example, Markov chain transition 
matrix is called ergodic, if all eigenvalues are below one, implying 
any state can be reachable from any other \cite{suezen2014a}. Here, 
spectral ergodicity implies eigenvalue spectra averaged over an ensemble 
of matrices and spectra obtained using a single matrix, a realisation 
from an ensemble, as it is or via an averaging procedure, 
i.e., {\it spectral average} generates the same spectra within 
the statistical accuracy.  It is known that, {\it spectral ergodicity} 
plays a vital role in interpreting neutron scattering experiments
\cite{jackson2001a}.

In the context of neural networks {\it spectral ergodicity} could manifest
in different ways. For example, in considering ensembles of weight matrices 
in network layers for feed-forward architectures, or the entire network 
architecture for recurrent networks. We conjecture that quantifying spectral 
ergodicity for these architectures would play an important role in understanding 
the peculiarities of training these networks and interpreting how they 
achieve high accuracy in learning tasks. To our best 
knowledge, nobody has so far been proposed a method to quantify 
spectral ergodicity aimed at deep learning architectures in a generic fashion. 
It was hinted that increasing ergodicity improves training accuracy of 
Restricted Boltzmann Machines \cite{desjardins10a}, but not spectral ergodicity. 
Our proposed metric, {\it spectral ergodicity} of weight matrices, is a 
generic approach without the need of accessing details of the learning 
algorithm or the specifics of the connection structure in the neural network. 

In this work we propose the following new approach to quantify 
spectral ergodicity for a {\it finite ensemble of random matrices}. 
First we define the {\it metric} 
\begin{equation} \label{equ1}
  \Omega^{N}_k \equiv \Omega^{N}(b_{k}) = 
  \frac{1}{M \cdot N} \sum_{j=1}^{M} \left[ 
                                           \rho_{j}(b_{k})-
                                           \bar{\rho}(b_{k})
                                     \right]^{2},
\end{equation}
for an ensemble of $M$ squared random matrices of size $N \times N$, 
where  $b_{k}$ with  $k=1,...,K$  are the discrete bins, $K$ is the 
number of bins and  $\rho_{j}(b_{k})$ is the spectral density of the 
$j-th$ matrix of the ensemble. The ensemble average spectral density is 
defined as 
\begin{equation} \label{equ2}
 \bar{\rho}(b_{k}) = \frac{1}{M} \sum_{j=1}^{M} \rho_{j}(b_{k}).
\end{equation}
As one can easily observe, this measure is inspired by the 
Thirumalai-Mountain (TM) metric structure \cite{mountain89me, 
 thirumalai1989ergodic}, although one should notice that in difference 
to the TM metric we apply it here not to a time-dependent observable but to 
the spectral density of the different matrices and the matrix size as 
the arguments of the new metric.  The essential aim of these new 
{\it metric} is to capture fluctuations between individual eigenvalue 
spectrum, i.e., which is a {\it spectral average} against the 
finite {\it ensemble average} one empirically. In the case of matrix 
size approaching very large values, the value of $\Omega^{N}_k$ should 
approach to zero.

\begin{figure}[ptb]
  \centering
  \subfigure[\label{fig:eigen_cse}]
             {
              \includegraphics[width=0.49\columnwidth]{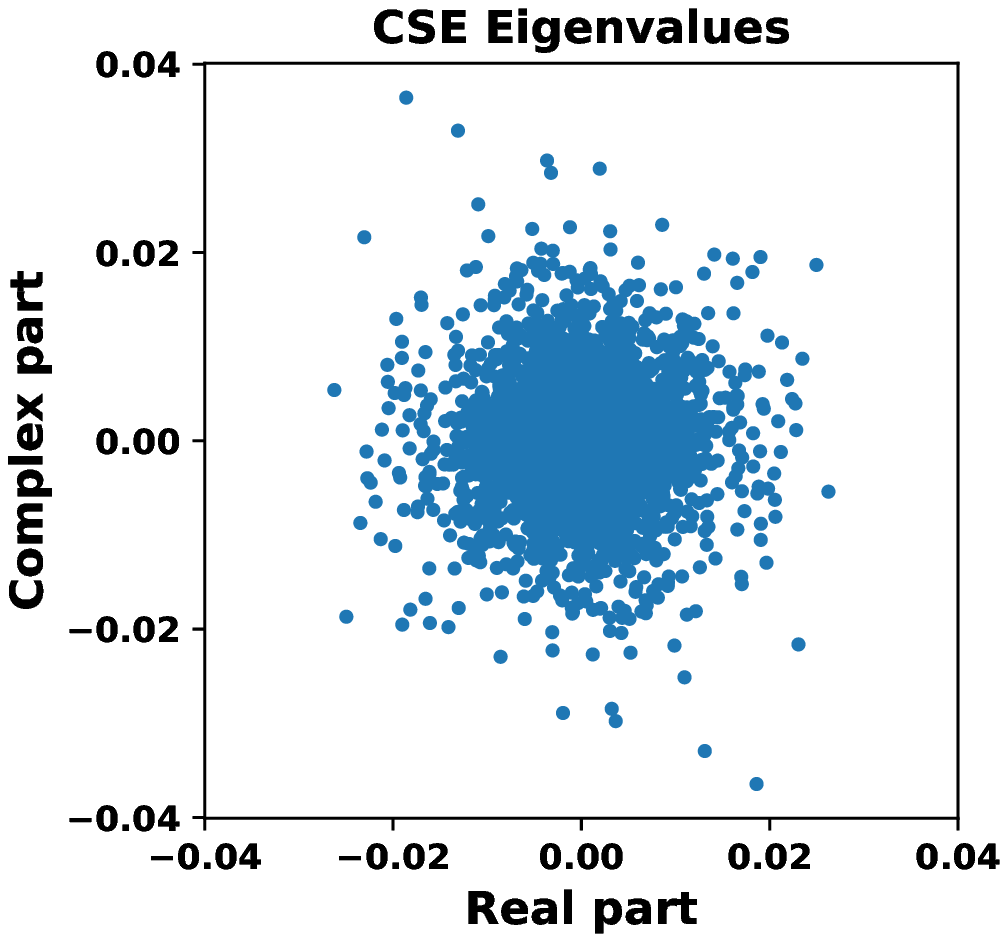}
             }
  \subfigure[\label{fig:eigen_cue}]
             {
              \includegraphics[width=0.47\columnwidth]{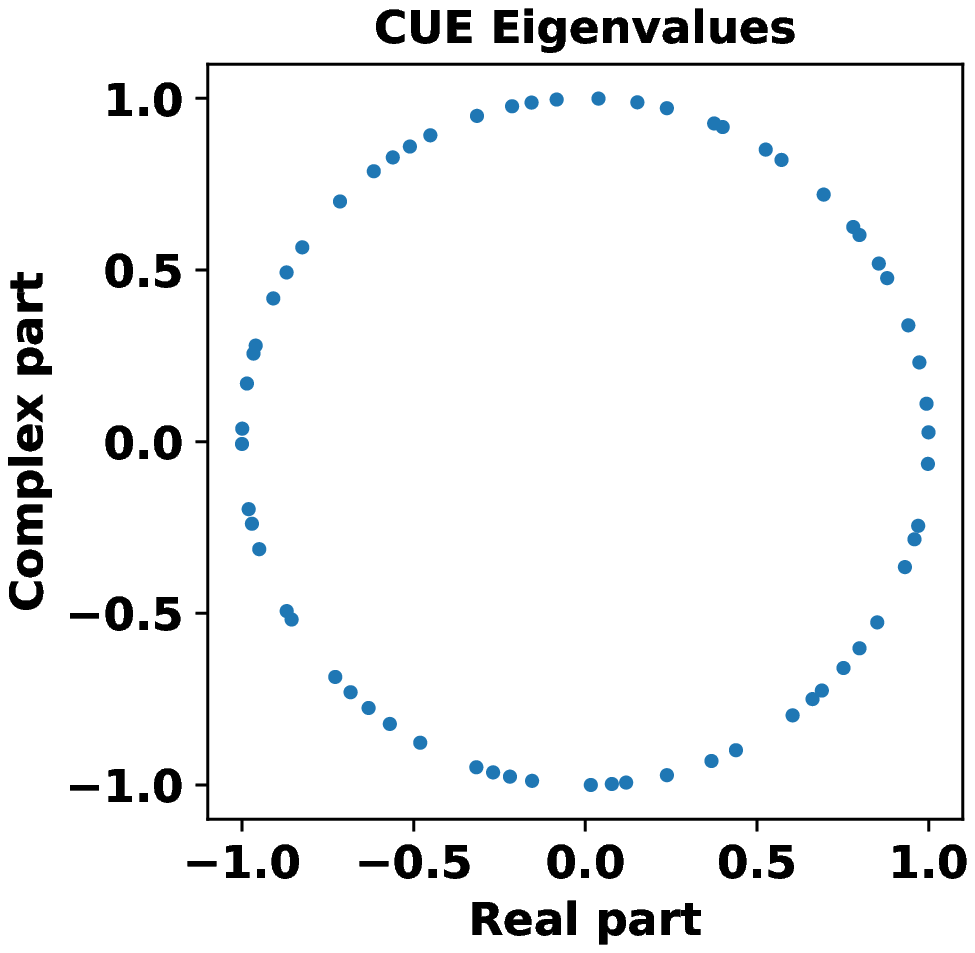}
             }
  \subfigure[\label{fig:eigen_coe}]
             {
              \includegraphics[width=0.47\columnwidth]{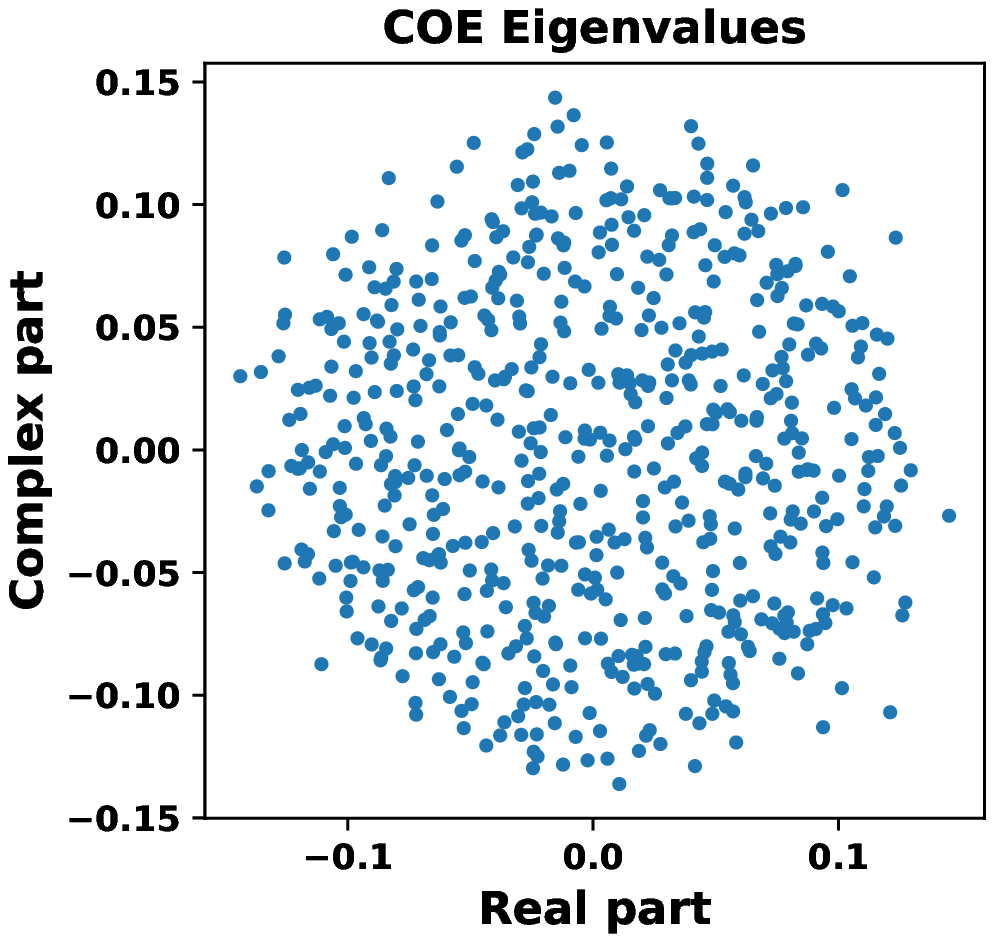}
             }
  \caption{An example set of eigenvalues on the complex plane for
           (a) CSE (b) CUE and (c) COE.}
\end{figure}

A second step is to define a distance metric  between two $\Omega$ 
distributions  $\Omega^{N_a}$ and $\Omega^{N_b}$ generated according to 
Eq. \ref{equ1} and corresponding to ensembles of matrix sizes $N_{a}$ and 
$N_{b}$, respectively. We define the distance metric as
\begin{equation} \label{equ3}
D_{se} (N_{a}, N_{b}) = D_{KL}(\Omega^{N_{a}} || \Omega^{N_{b}}) 
               + D_{KL}(\Omega^{N_{b}} || \Omega^{N_{a}}),
\end{equation}
where 
\begin{equation} \label{equ4}
D_{KL}(\Omega^{N_{a}} || \Omega^{N_{b}}) =  \sum_{k=1}^{K} \Omega^{N_{a}}_{k}
                                            \log_{2}(\Omega^{N_{a}}_{k}/
                                                     \Omega^{N_{b}}_{k}).
\end{equation}
The sum of two terms in Eq. (\ref{equ3}) is due to the non-symetric nature 
of Eq. (\ref{equ4}). Clearly, the new metric based on Eqs. (\ref{equ3}) 
and (\ref{equ4}) is rooted on the Kullbach-Leibler (KL) divergence 
\cite{kullback1951a, bishop2007a}, although one should notice that we do not 
apply the new metric to probability distribution functions but to $\Omega$ 
distributions generated according to Eq. \ref{equ1}. $D_{se} (N_{a}, N_{b})$ 
allows to assess how increasing network size, i.e. larger connectivity 
matrix, influences the approach to spectral ergodicity.  In order to test 
our new method in the field of deep learning,  instead of working on a 
specific architecture and learning algorithm, we use random matrices to 
mimic a generic ensemble of weight matrices, such as in feed-forward 
architectures, i.e., deep learning \cite{lecun2015a,bengio2009a} or in
recurrent networks \cite{lukovsevivcius09a}. Using 
random matrices brings a distinct advantage of generalising the results for 
any architecture using a simple toy system and being able to enforce certain 
restrictions on the spectral radius. For example, applying a constraint on 
the spectral radius to unity, this simulates preventing vanishing or exploding
gradient problems in training the network \cite{pascanu13a}. Circular random matrix ensembles 
serve this purpose well, where their eigenvalues lie on the unit circle in 
the complex plane. A typical weight matrix for a layer, $W$, in deep 
feed-forward multi-layer neural networks, is not square. In those cases one 
can proceed with the square matrix $W^T W$ to find the eigenvalue spectrum.

Implementation of parallel code generating circular random matrix 
ensembles \cite{bristol_py, dataset}, using Mezzadri's approach 
\cite{mezzadri06a,berry2013hearing}, is utilized. Using circular ensembles 
circumvents the need for spectral unfolding \cite{haake2013a}. 
Random seeds are preserved for each chunk from an ensemble, so that 
the results are reproducible both in parallel and serial runs 
\cite{bristol_py}. Three different circular ensembles are generated. 

Realisations for circular ensembles can be generated as follows. 
Consider a Hermitian matrix $H \in \mathbb{C}^{N \times N}$, in 
component form, $$H_{ij} = \frac{1}{2}(a_{ij}+ I b_{ij}+a_{ji}- I b_{ji}),$$
where $1 \leq i,j \leq N$, and $a_{ij}, b_{ij}, 
a_{ji}, b_{ji} \in \mathbb{G}$, i.e, they are elements of 
the set of independent identical distributed Gaussian random 
numbers sampled from a normal distribution and $I$ is 
the imaginary number.

The Circular Unitary Ensemble (CUE) is defined as
$$U = exp(\gamma_{i} I) \cdot v_{j}^{i},$$
$v_{i}$ is the $i$-th eigenvector of $H_{ij}$, 
where $\gamma_{i} \in [0, 2\pi]$ is a uniform random number. 
The Circular Orthogonal Ensemble (COE) is dependent on CUE as follows,
$$O = U^{T} U.$$
The Circular Symplectic Ensemble (CSE) also depends on CUE,
$$S = (Z U^{T} Z) U,$$
where $S \in \mathbb{C}^{2N \times 2N}$, $Z \in \mathbb{R}^{2N \times 2N}$
a symplectic matrix obtained by the outer product of the $N \times N$ 
identity matrix with the unit antisymmetric $2 \times 2$ matrix. In summary,
$Z$ is formed by placing blocks of antisymmetric unit matrices of size $2 \times 2$
over diagonals of  $2N \times 2N$ matrix, and leaving off-diagonal 
elements zero.

Simulated circular ensembles consist of matrices of different sizes: 
$64, 128, 256, 512, 768, 1024$ with ensemble size of $40$ each. 
Eigenvalues of all matrices lie within a unit circle on the complex plane.  
We see that typical eigenvalues for CSE are concentrated in a more 
dense region in Figure \ref{fig:eigen_cse}. 
Similarly, typical eigenvalues are shown in Figure \ref{fig:eigen_cue} 
and Figure \ref{fig:eigen_coe}, for CUE and COE respectively. 
They are more spread for COE. In the case of CUE, all eigenvalues 
lie on the unit circle.

\begin{figure}[ptb]
  \centering
  \includegraphics[width=0.9\columnwidth]{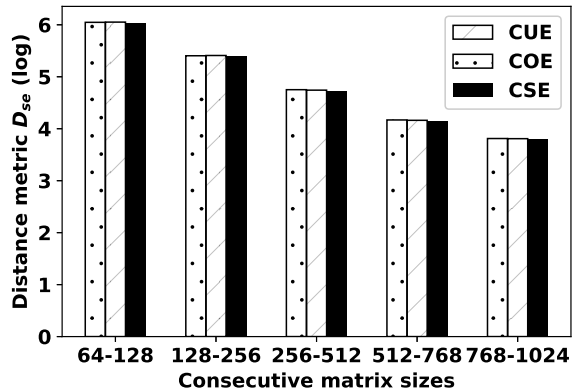}
  \caption{Distance metric for spectral ergodicity, $D_{se}(N_{j})$, 
           for CUE, COE and CSE. 
           Labels shows the consecutive sizes 
           used in computing $D_{se}(N_{a})$ on the horizontal axis.}
  \label{fig:distance_Dse}
\end{figure}

Approach to spectral ergodicity $D_{se} (N_{a}, N_{b})$ for all circular 
ensembles is computed. Spectra are constructed via a histogram of 
arguments of eigenvalues, as they are complex, 
with a fixed binwidth over all ensembles. It is seen that 
results are robust against different binwidths. 

As in Figure \ref{fig:distance_Dse} where $D_{se} (N_{a}, N_{b})$ distance 
is shown as a function of the matrix size, the surrogate weight matrices 
at fixed size of the ensembles, a consistent decrease in 
$D_{se}(N_{a}, N_{b})$ with increasing matrix size is observed. 
This can be interpreted as {\it spectral ergodicity} being a property 
of deep neural networks; an ensemble here can be thought as the number 
of layers having fixed number of neurons at each layer, matrix size 
of $N$. With this fixed ensemble size, we observe that using larger 
sizes of hidden layers leads to spectral ergodicity faster.  

Hence, $D_{se} (N_{a}, N_{b})$ can help us to identify a lower bound on 
how large a layer to use in learning algorithms. One could identify 
an ensemble of architectures or different depths of layers, and construct 
an ensemble of weight matrices.  An optimal combination of learning 
algorithm and architecture can be identified when {\it spectral ergodicity} 
is reached within a given threshold. This approach would save the 
practitioner both computation time and design effort.

COE and CSE ensembles produce smaller eigenvalues, with a spectral 
radius $\rho$, less than  0.2. This may generate a vanishing gradient issue in
learning algorithms \cite{pascanu13a}, however the elements, i.e. weights of
COE and CSE matrices can be upscaled so that the maximal eigenvalue is 1,
this numerical change will not change the generic behavior based on 
{\it spectral ergodicity} we have observed.
 
Using information-theoretic measures to understand deep learning has been
recently explored \cite{tishby2015a}. In this framework, it is argued 
that the optimal architecture, number of layers and connection in each 
layer, can be determined via propagation of the mutual information (MI) 
between the layers. Even though, our new metric $D_{se} (N_{a}, N_{b})$ 
is not formally a Kullbach-Leibler distance, so it is not an information 
metric, we can capture similar characteristics from the layers without 
the need of prior knowledge about the learning algorithm or the training 
data.  Our approach only requires weight matrices.

Success of deep learning architectures is attributed to availability of 
large amounts of data and being able to train multiple layers at the same 
time \cite{lecun2015a,bengio2009a}. However, how this is possible from a 
theoretical point of view is not well established. We introduce quantification 
of spectral ergodicity for random matrices as a surrogate to weight matrices 
in deep learning architectures and argue that spectral ergodicity 
conceptually can improve our understanding about how these 
architectures perform learning with high accuracy. From a biological
 standpoint, our results also show that spectral ergodicity would play 
an important role in understanding synaptic matrices, as random matrices 
are used in understanding dynamics and memory in the brain as 
well \cite{rajan2006b, aljadeff15a, rajan2016a}.

We thank Christian Garbers for valuable discussion on circular statistics 
and code review. Ruben Foission and Marcel Reginatto for sharing their notes 
on spectral unfolding.

\bibliographystyle{apsrev4-1}
\bibliography{lib}
\end{document}